\documentclass[journal]{IEEEtran}
\usepackage{times}
\usepackage{epsfig}
\usepackage{graphicx}
\usepackage{amsmath}
\usepackage{amsthm}
\usepackage{amssymb}
\usepackage{booktabs}
\usepackage{bm}  
\usepackage{algorithm}
\usepackage{algorithmic}
\usepackage{mathtools}
\usepackage{enumerate}
\usepackage{paralist}
\usepackage{subcaption}
\usepackage{color}
\usepackage{cite}
\usepackage{multirow}
\usepackage{array}

\usepackage{mathtools}
\usepackage{amsmath}
\usepackage{amssymb}
\usepackage{amsthm}
\usepackage{color}
\usepackage{bm}
\usepackage{booktabs}
\usepackage{multirow}
\usepackage{arydshln}
\usepackage{enumitem}

\newcommand{\PathFig}{./fig}   

\newcommand{\matx}[1]{{\bm{#1}}} 
\newcommand{\vect}[1]{{\bm{#1}}} 
\newcommand{\set}[1]{\mathbb{#1}} 
\newcommand{\opt}[1]{\mathcal{#1}} 
\newcommand{\va}{\vect{a}}

\newcommand{\vd}{\vect{d}}

\newcommand{\vf}{\vect{f}}
\newcommand{\vg}{\vect{g}}

\newcommand{\vn}{\vect{n}}

\newcommand{\vr}{\vect{r}}
\newcommand{\vs}{\vect{s}}

\newcommand{\vu}{\vect{u}}

\newcommand{\vw}{\vect{w}}

\newcommand{\mA}{\matx{A}}

\newcommand{\mG}{\matx{G}}

\newcommand{\mR}{\matx{R}}

\newcommand{\mV}{\matx{V}}

\newcommand{\sA}{\set{A}}

\newcommand{\sE}{\set{E}}

\newcommand{\sN}{\set{N}}

\newcommand{\sR}{\set{R}}
\newcommand{\sS}{\set{S}}

\newcommand{\oN}{\opt{N}}

\newcommand{\ie}{{\it i.e.}}
\newcommand{\eg}{{\it e.g.}}

\newcommand{\etal}{{\it et al.}}

\DeclareMathOperator*{\argmax}{argmax}

\newtheorem{remark}{Remark}[]

\renewcommand{\div}{{\rm div}}
\newcommand{\norm}[1]{\lVert#1\rVert}
\newcommand{\XRT}[1]{\textcolor[rgb]{0.00,0.00,0.00}{#1}}   

\begin{document}

\title{Reinforced Diffusion: Learning to Push the Limits of Anisotropic Diffusion for Image Denoising}
\author{
	Xinran Qin,Yuhui Quan,  Ruotao Xu and Hui Ji
	\thanks{Xinran Qin, Yuhui Quan, Ruotao Xu are with School of Computer Science and Engineering at South China University of Technology, Guangzhou 510006, China.}
	\thanks{Hui Ji is with Department of Mathematics at National University of Singapore, Singapore 119076.}

}

\markboth{IEEE Transcations on XXXXX}{}
\maketitle

\begin{abstract}
Image denoising is an important problem in low-level vision and serves as a critical module for many image recovery tasks. 
Anisotropic diffusion is a wide family of image denoising approaches with promising performance. However, traditional anisotropic diffusion approaches use explicit diffusion operators which are not well adapted to complex image structures. As a result, their performance is limited compared to recent learning-based approaches. In this work, we describe a trainable anisotropic diffusion framework based on reinforcement learning. By modeling the denoising process as a series of naive diffusion actions with order learned by deep Q-learning, we propose an effective diffusion-based image denoiser. The diffusion actions selected by deep Q-learning at different iterations indeed composite a stochastic anisotropic diffusion process with strong adaptivity to different image structures, which enjoys improvement over the traditional ones. The proposed denoiser is applied to removing three types of often-seen noise. The experiments show that it outperforms existing diffusion-based methods and competes with the representative deep CNN-based methods.
\end{abstract}

%
\IEEEpeerreviewmaketitle


\section{Introduction}\label{sec:intro}
It often happens that digital images contain noticeable amount of measurement noise due to the interference from imaging equipment and external environment. The noise not only has negative visual effects, but also lowers the effectiveness of many image processing and recognition algorithms. 
Let $\Omega\subset\sR^2$ denote a subset of the plane, and $\vg,\vf:\Omega\rightarrow\sR$ denote the observed noisy image and its latent clean image respectively. The formation of $\vg$ can be formulated as follows:
\begin{equation}
\vg=\vf+\vn,
\end{equation}
where $\vn:\Omega\rightarrow\sR$ denotes the measurement noise.
Image denoising refers to recovering $\vf$ from $\vg$. It is not only an important task in computer vision with great practical vaules, but also a basic tool for many image recovery tasks. 

\subsection{Diffusion-Based Image Denoising}
Partial differential equations (PDEs) have become one often-used approach for various problems in image processing. They come along with a sound mathematical framework that allow to make clear statements about the existence and regularity of the solutions. 
Let $\vu^{(t)}:\Omega\rightarrow\sR$ be a family of gray-scale images. In general, the PDE-based approaches to image denoising model the denoising process as a diffusion process in the following form:
\begin{equation}\label{eq:naive}
	\left\{
		\begin{array}{l}
			\frac{\partial \vu}{\partial t}={\div}(\vd\XRT{(x,y)}\nabla\vu)
			\\
			\vu^{(0)}=\vg
		\end{array}
	\right.,
\end{equation}
where $\nabla$, ${\div}$ denote the {gradient} and divergence operator respectively,
and the diffusivity function $\vd:\Omega\rightarrow\sR$ controls the patterns of diffusion such as amount and direction of diffusion at each location. It is well-known that when $\vd$ is set to constant, the diffusion is performed homogeneously in all the directions and locations, which is equivalent to a naive isotropic Gaussian filtering that tends to blur image edges.

For better edge preservation, Perona and Malik~\cite{perona1990scale} proposed the anisotropic diffusion, which defines $\vd(x,y)=h(\norm{\nabla\vu(x,y)})$ with some monotonic decreasing function $h:\sR\rightarrow\sR$, \eg~$h(z)=1/(1+z^2)$. This results in  the so-called anisotropic diffusion which applies different diffusion at different image locations.


\XRT{To further free the user from the difficulty of specifying an appropriate stopping time for the diffusion process, the variants (\eg~\cite{nordstrom1990biased}) known as reaction diffusion were proposed by introducing a bias term, which reacts against the strict smoothing effect of the pure anisotropic diffusion.}

\subsection{Motivations and Main Idea}
Although recent PDE approaches have shown good performance for a number of image processing tasks, they still fail to produce state-of-the-art quality for classical image restoration tasks such as image denoising.
A proper definition of the diffusion process is the key to the success of the diffusion-based denoiser. Two kinds of decisions are involved in the diffusion process:
\begin{enumerate}
	\item Where and how much should the energy/intensity diffuse? Diffusion should happen in a homogeneous region to remove the noises, while there should not be diffusion across the region boundaries or edges for edge preservation.
	\item When should the diffusion stop? Diffusion should stop once the noise is eliminated. Excessive steps of diffusion will lead to over-smoothness.
\end{enumerate}	
Note that the decisions should vary in positions, as pixels locate on an edge or in a homogeneous region should be treated differently. 
The decisions should also vary in steps. Indeed, these decisions are  made depended on the locations of edge  or amount of remaining noises, which is usually estimated from the image at the current step, \eg~$h(\norm{\nabla\vu^{(t-1)}})$ in the classic anisotropic diffusion. 
In the classic anisotropic diffusion model~\cite{perona1990scale}, all these decisions are implied by the diffusivity function $\vd$, whose design is generally a difficult task, as 
good insights into the task and a deep understanding of the behavior of the PDEs are usually required.

Instead of using a hand-crafted diffusivity function $\vd$ to make the decisions, in this paper, we proposed to learn the decision process in a data-driven way.
Considering the decision process usually satisfy the Markov property, \ie~the decisions are determined by the generated image at the current step, 
we first formulated the diffusion process as a Markov Decision Process (MDP), where the state includes current images, the action determines and how to diffuse and whether to stop for each pixel and the reward depends on the recovery quality. 
Note that the diffusion of each step can be view as a weighted average among neighbors for each pixel, which can be represented as composition of several pair-wise averages with neighbors. We define a simple action set with $9$ actions: $8$ pair-wise average with $8$ different neighbor pixels and $1$ "do nothing" for the stopping decisions.
We further solve the MDP with a multi-agent deep reinforcement learning framework, where pixel-wise agents are supervised learned to make spatially-varying decisions. Benefiting from the simple action set, the deep reinforcement learning process is stable and able to converge in \XRT{a few iterations}. With the learned agents, the diffusion in each time-step is performed according to the decision of each agent to remove the noises. 
The proposed method shows comparative results with existing ones in different denoising tasks, including Gaussian denoising, Salt and Pepper noise removal and Poisson denoising.

\subsection{Contributions}
The contributions of this paper is summarized as follows:
\begin{itemize}
	\item We propose a novel learning-based anisotropic diffusion model for image denoising. The traditional diffusion models are  usually hand-crafted, which need heavy expert knowledges to the task as well as the PDE theories. In contrast, the proposed model learns the diffusion process in a data-driven way, which can be easily adapted to different denoising tasks. The proposed method achieves a large gap of improvements over the existing diffusion-based method, which has pushed the limits of anisotropic diffusion for image denoising.
	\item We model the diffusion process as a Markov decision process with $9$ actions in the action sets. The proposed actions set, though simple, can compose complex anisotropic diffusion in several diffusion iterations, which leads to a content-aware anisotropic-diffusion-based image denoiser.
	\item We \XRT{solve} the Markov decision process with a multi-agent deep reinforcement learning framework, where pixel-wise agents are learned to decide where to diffuse and whether to stop for each pixel in each step.  To the best of our knowledge, the proposed model is the first one that learns the diffusion model with deep reinforcement learning framework \XRT{with agent-per-pixel configuration}.
\end{itemize}

\section{Related work}\label{sec:bg}
\textbf{Image denoising.}
There is abundant literature on image denoising. The very early approaches modeled image denoising as a filtering problem or a diffusion process. 
See~\cite{zhang2012effective} for a comprehensive survey on the filtering-based methods.
Regarding the diffusion-based method, the early work can be traced back to Iijima's work~\cite{iijima1962basic} in 1962, which proposes the linear diffusion model. Due to the isotropic diffusion nature, this method is known to suffer from the problem of over-smoothness. To overcome this problem, in another seminal work~\cite{perona1990scale}, Perona and Malik (PM) propose a nonlinear diffusion model, which controls the strength of diffusion of different pixels with a so-called edge-stopping function~\cite{black1997robust} or diffusivity function~\cite{weickert1998anisotropic}. The proposed PM diffusion model  leads to a nonlinear anisotropic diffusion model which is able to preserve and enhance the image edges. Though with the edge-stopping function, the PM model would still blur the edges as the iteration grows. Hence, it is critical to choose a proper stopping time or criterion for different images. To free the user from the difficulty of specifying an appropriate stopping time, a variant of the PM model called reaction diffusion~\cite{nordstrom1990biased} is proposed, which introduces an additional term to reacts against the strict smoothing effect of the PM diffusion, leading to a non-trivial steady-state. Subsequent works consider more complicated diffusion processes~\cite{gilboa2002forward,welk2009theoretical,didas2009properties,guidotti2011two,hajiaboli2011anisotropic} or reaction terms~\cite{acton2001oriented,cottet1993image,esclarin1997image} for further improvements.
It should be noted that the diffusion models mentioned above are all handcrafted, including elaborated selections of the edge-stopping functions, optimal stopping times and proper reaction forces, which requires heavy expert knowledge to the behavior of PDEs. 

Instead of directly modeling the denoising process, some methods solve the denoising problem by exploiting certain image priors to overcome the undeterminedness. Considerable efforts have been made in designing the proper image priors.  In the last two decades, sparsity prior have become one preferred choice for denoising, which assumes  a noise-free image should be sparse under certain transforms, \eg~wavelet transform~\cite{chen2005image}, curvelet transform~\cite{starck2002curvelet},rigdelet transform~\cite{chen2007image} or adaptive dictionaries~\cite{hou2003adaptive,elad2006image,bao2015dictionary}. Another prominent prior is the nonlocal self-similarity(NNS) prior, which makes use of the recurrent patches within an image. BM3D~\cite{dabov2008image} is arguably the most popular one which applies collaborative filtering to similar patches. WNNM~\cite{gu2017weighted} and TWSC~\cite{xu2018trilateral} are another two popular self-similarity-based  methods that impose the low rank property on similar pathces for denoising. 

Instead of using the sparsity prior or the patch recurrence prior, some approaches learn image priors from visual data. Portilla~\etal~\cite{portilla2003image} proposed to learn a Gaussian scale mixture model on the wavelet coefficients of natural images. Roth~\etal~\cite{roth2009fields} proposed to learn a high-order Markov random field for modeling natural images. The classic EPLL approaches~\cite{zoran2011learning} learns a Gaussian mixture model of image patches. Xu~\etal~\cite{xu2015patch} proposed to learn the distribution prior on similar patch groups. 
Instead of learning image priors, an alternative approach is to directly learn the denoising process. 
Schmidt~\etal~\cite{schmidt2014shrinkage} models image denoising as a cascade of shrinkage fields with learnable parameters of filters and shrinkages. 
Chen~\etal~\cite{chen2017trainable} propose a learnable diffusion-based denoiser called Trainable Nonlinear Reaction Diffusion (TNRD),  which formulates the PM model in a discrete and decouple form with trainable convolutional filters and influence functions. Following TNRD, multi-scale scheme~\cite{feng2016image} and non-local mechanism~\cite{qiao2017learning} are also incorporated in TNRD to further boost the denoising performance. Unlike TNRD and its extensions, which learn a universal diffusion model for all images, we propose a content-aware diffusion model, whose diffusion process are determined according to the input images and the intermediate denoised images.

More recently, neural-network-based methods achieved remarkable performance. An early work with neural networks~\cite{burger2012image} used a multilayer perceptron discriminatingly trained on synthetic Gaussian noise and showed significant improvements. Zhang~\etal~\cite{zhang2017beyond} proposed DnCNN, which showed that the residual structure and the use of batch normalization can greatly helps the denoising tasks. Following DnCNN, many other architectures have been proposed, such as RED~\cite{mao2016image}, MemNet~\cite{tai2017memnet}.  Due to the local nature of the convolution operator, the CNN-based methods usually have a regular and limited receptive field for each neuron-pixel. As a result, they are unable to  exploit the self-similar patterns that were proven to be highly successful in denoising tasks. To address this issue, a few works~\cite{cruz2018nonlocalityreinforced,lefkimmiatis2018universal,plotz2018neural,liu2018non,valsesia2020deep,Quan2024,Qin2025,Qin2023,Qin2025a,Zhang2023,Zhang2024} tried to incorporate the nonlocal information into CNNs by adding a nonlocal means filter or a BM3D denoiser in or after the CNN-based denoisers. Note that all the NN-based denoisers mentioned above, though with different architectures, such as convolution layer, recurrent layer~\cite{liu2018non} and graph convolution layer~\cite{valsesia2020deep},  attempt to learn the mapping from noisy image to clear image directly. Unlike these methods, the proposed method learns CNNs to determine denoising processing in a deep reinforcement framework, which is able generate spatial-variant filters with spatial-variant shape of receptive field for image denoising.

\textbf{Deep Reinforcement Learning for image restoration.}
Reinforcement learning is a process in which an agent learns to make decisions through trial and error, which is often modeled mathematically as a Markov decision process(MDP). Deep reinforcement learning(DRL) algorithms~\cite{mnih2013playing,mnih2016asynchronous,schulman2015trust,nachum2017bridging} incorporate deep neural network to solve MDPs by representing the policy or the learned functions as neural networks.
The primary application of DRL is in playing video games such as Atari games.

Recently, DRL has also been applied to some image restoration applications. Cao et al.~\cite{cao2017attention} proposed a super-resolution method for face images, where the agent chooses a local region and inputs it into a trainable super resolution network in each step. Yuan et al.~\cite{yu2018crafting} proposed a DRL method for image restoration, where the agent selects a toolchain from a set of light-weight CNN to restore a corrupted image. Note that both methods mentioned above introduce a single agent and can execute only global actions for the entire image. 
Different from these methods, Furuta et al.~\cite{furuta2020pixelrl} proposed a agent-per-pixel DRL algorithm called PixelRL for image denoising and restoration, whose pre-dinfed action set includes different filtering operator, pixel value increment/decrement, and "do nothing". Following~\cite{furuta2020pixelrl}, Vassilo et al.~\cite{vassilo2020multistep} proposed a DRL method for single image super-resolution with agent-per-patch configuration, where each agent chooses an action from a fixed actions set comprised of results from existing GAN SISR  algorithms. However, there is few attempts to learn a diffusion model with a deep reinforcement framework. In this paper, we show that a diffusion process can be naturally modeled as an MDP and a agent-per-pixel deep reinforcement network to make spatial-variant decisions for each diffusion steps.  As far as the authors know, the proposed method is the first one to learn a diffusion model with a deep reinforcement framework.
\section{Preliminaries}\label{sec:A3C}
A sequential decision problem can be formulated as a Markov Decision Process (MDP). MDP is a tuple $ (\sS,\sA,\tau,r,\gamma) $ that consists of the state space $ \sS $, the action space $ \sA $, 
the probabilistic transition function of the environment $ \tau:\sS\times\sA\times\sS\rightarrow [0,1]$ 
, the reward $r: \sS\rightarrow\sR^{|\sA|}$ and the discount factor $\gamma$. A policy $\pi$ in RL is a probability distribution on the action $\sA$ over $\sS$: $\pi:\sS\times\sA \rightarrow [0,1]$. 
The return $G^{(t)}$ is the total accumulated return from time step $t$ with discount factor $\gamma$:
\begin{equation}\label{eq:return}
	G^{(t)}=r^{(t)}+\gamma G^{(t+1)}=\sum_{k=0}^\infty \gamma^kr^{(t+k)}.
\end{equation}
Given a policy $\pi$, the value of state $s$ is defined as 
$$
V^\pi(s)=\sE_\pi[G^{(t)}|s^{(t)}=s],
$$
which is simply the expected return for following policy $\pi$ from state $s$.

Given an MDP, the goal of a reinforcement learning algorithm is to find a policy $\pi$ that maximizes the discounted accumlated rewards from the initial state in this MDP:
$$
	\pi^*=\argmax_\pi~\sE_\pi[G^{(0)}]=\sE_\pi[\sum_{t=0}^\infty\gamma^tr^{(t)}].
$$

Many effective RL algorithms have been developed to find the optimal policy $\pi^*$. 
In this paper, we extend the asynchronous advantage actor-critic (A3C)~\cite{mnih2016asynchronous} for the learning-based diffusion model because A3C showed good performance with efficient training in the original paper. Here, we give a brief review on the training algorithm of A3C. A3C is one of the actor-critic methods, which has two networks: policy network and value network. We denote the parameters of each network as $\theta$ and $\rho$, respectively. Both networks use the current state $s^{(t)}$ as the input. The value network aims to approximate the state value function: $V_\theta(s)\approx V^{\pi^*}(s)$. In one-step learning, the parameters $\theta$  of the value network $V_\theta(s)$ are learned by iteratively minimizing a sequence of loss functions.
%
The loss function at time $t$ is defined as
$$
\ell(\theta)=\sE[r^{(t)}+\gamma V_{\theta'}(s^{(t+1)})-V_\theta(s^{(t)})]^2
$$
where $\theta'$ is the parameters before update. The gradient for $\theta$ can then be computed as 
$$
d\theta=\nabla_{\theta}(r^{(t)}+\gamma V_{\theta'}(s^{(t+1)})-V_\theta(s^{(t)}))^2,
$$

The policy network aims to approximate the optimal policy: $\pi_\rho(a|s)\approx\pi^*(a|s)$. Therefore, the output dimension of the policy network $\pi_\rho(a|s)$ is $|\sA|$. In A3C, the parameters $\rho$ are learned by iteratively maximizing the expected advantages. The expected advantage at time $t$ is defined as:
$$
	\sE_{\pi_{\rho}}[A(a^{(t)}|s^{(t)})]=\sum_{a^{(t)}\in\sA}\pi_\rho(a^{(t)}|s^{(t)})A(a^{(t)}|s^{(t)})
$$
where the advantage $A(a^{(t)}|s^{(t)})$ is defined as
$$
	A(a^{(t)}|s^{(t)})=r(s^{(t)},a^{(t)})+\gamma V_\theta(s^{(t+1)})-V_\theta(s^{(t)}).
$$
Note that $V_\theta(s)$ is subtracted to reduce the variance of the gradient. The gradient of $\rho$ can then be computed as follows:
$$
d\rho=-\nabla_{\theta_p} \log \pi(a^{(t)}|s^{(t)})A(a^{(t)}|s^{(t)}).
$$
For A3C, the parameters are trained in parallel, where multiple workers in parallel environments independently update the global networks, leading to the asynchronous property. For more details, see~\cite{mnih2016asynchronous}.

\section{MDP modeling for Diffusion Model}\label{sec:MDP}
Given the diffusivity function $\vd^{(t)}$, the diffusion~\eqref{eq:naive} can be performed iteratively as follows:
\begin{equation}\label{eq:diffusion}
	\begin{array}{ll}
		\vu^{(t+1)}&=\vu^{(t)}+\kappa\div(\vd^{(t)}\nabla\vu^{(t)})\\
		&=\vu^{(t)}+\kappa\nabla\vd^{(t)}\cdot\nabla\vu^{(t)}+\kappa\vd^{(t)}\Delta\vu^{(t)},
	\end{array}
\end{equation}
where $\nabla$, $\Delta$ and $\cdot$ denote the gradient operator, laplacian operator and the inner product, respectively. When using discrete implementation s of the gradient and laplacian operators
, the diffusion process  \eqref{eq:diffusion} is equivalent to a weighted summation of the adjacent pixel values. For convenience, let $u_{x,y}=\vu(x,y)$ and $d_{x,y}=\vd(x,y)$ with $(x,y)\in\Omega$, then \eqref{eq:diffusion} can be rewritten in the following form:
\begin{equation}\label{eq:summation}
\begin{array}{lll}
	u^{(t+1)}_{x,y}
	&=&\sum_{(i,j)\in\sN}w^{(t)}_{x,y,i,j}u_{x+i,y+j}^{(t)}
\end{array}	
\end{equation}
where $\sN=\{0,-1,1\}^2$, and
\begin{equation}\label{eq:weights}
w^{(t)}_{x,y,i,j}=\left\{
	\begin{array}{ll}
		1-\kappa d^{(t)}_{x,y},& i=0,j=0,\\		
		\frac{\kappa d^{(t)}_{x,y}}{8}+\frac{i\nabla_x\vd^{t}\kappa d^{(t)}_{x+i,y+j}}{6}\\
		+\frac{j\nabla_y\vd^{(t)}\kappa d^{(t)}_{x+i,y+j}}{6},& \text{otherwise}.\\
	\end{array}
\right.
\end{equation}

When $\vd^{(t)}$ is a constant function,  Eq~\eqref{eq:summation} is simply an isotropic average which tends to blur the edges. In the existing anisotropic diffusion model, the  diffusivity functions $\vd^{(t)}$ are carefully designed, and the  weights $w^{(t)}$ can then be implied to perform an anisotropic weighted average that leads to edge-preserving effects. 

Unlike these methods, in this paper, we tend to directly learn the optimal weighted averages for each pixel in each iteration. Note that the diffusion process can be naturally formulated as a sequential decision process with multi agents, where each agent find and perform the optimal weighted average for the corresponding pixel. Furthermore, such sequential decision process satisfy the Markov process in most cases. That is to say, at time step $t$, the optimal weighted average can be found depended on the currently generated images $\vu^{(t)}$, without the previous generated image $\vu^{(i)}$, $i<t$. Therefore, we can formulate the diffusion process as a Markov Decision Process~(MDP). 
The details of the MDP modeling for the diffusion process are elaborated as follows:

\textbf{States:} the state at time $t$ is denoted by $\vs^{(t)}:\Omega\rightarrow \sR$, which equals the generated image at time-step $t$
.
The initial state is set as $\vs^{(0)}=\vg$.

\textbf{Actions:} the action of the $(x,y)$-th agent at time $t$ is denoted by $a_{x,y}^{(t)}$, which performs the proper weighted summation over the adjacent pixels of the $(x,y)$-th pixel. We use the same action set $\sA$ for all the agents. With the definition of the actions for each agent, the action for the whole MDP at time $t$ can be defined as $\va^{(t)}: \Omega\rightarrow\sA$, which satisfied $\va^{(t)}(x,y)=a_{x,y}^{(t)}$. Note that size of the action space of $\va^{(t)}$ is $|\sA|^{|\Omega|}$, which is exponential to the size of each sub-actions $a_{x,y}^{(t)}$. It is generally difficult for reinforcement learning algorithm to explore a large action space. Therefore, it is critical to design a small but sufficient sub-action set $\sA$. The design of $\sA$ will be detailed in the next section.

\textbf{Transition}: we use a deterministic transition function in the MDP, which generates the new state $\vs^{(t+1)}$ by performing the $(x,y)$-th action $a_{x,y}^{(t)}$ on the each $(x,y)$-th pixel in  the current state $\vs^{(t)}$.

\textbf{Rewards}: as a deteministic transition is used, we can directly defined the reward function on the new state generated by the corresponding action. The reward of the $ (x,y) $-th agent at time $t$ is computed as $r_{x,y}^{(t)}=(f_{x,y}-u^{(t)}_{x,y})^2-(f_{x,y}-u_{x,y}^{(t)})^2$, where $f_{x,y}=\vf(x,y)$ is the value of the $ (x,y) $-th pixel in the groundtruth image. The rewards $r_{x,y}^{(t)}$s measure the descent of MSE errors. For the whole MDP, the rewards can be represented as a two-dimensional function $\mR^{(t)}: \Omega\rightarrow \sR$, where $\mR^{(t)}(x,y)=r^{(t)}_{x,y}$.

\textbf{Discount factor}: we set $\gamma=0.95$ for the discount factor in our implementation.

\section{Directional Action set}
As metioned above, a proper design of the sub-action space $\sA$ is critical to the MDP. Considering~\eqref{eq:summation}, an intuitive setting of the action space is as follows:
$$
	\{	u^{(t+1)}_{x,y}
	=\sum_{(i,j)\in\sN}w^{(t)}_{i,j}u_{x+i,y+j}^{(t)}|w_{i,j}\in\sR\},
$$
which is equivalent to $\sR^9$. Since we use a multi-agent MDP, the action space for the whole MDP is then $\sR^{9|\Omega|}$, which is extremely large. Such a continuous and high-dimensional space would make it hard for the agents to explore the action space, which is known to result in instability in deep reinforcement learning. In this paper, we will point out that a good choice of the direction of diffusion is \XRT{sufficient} for a diffusion-based denoiser . In other words, it is acceptable to choose a proper adjacent pixel for average without carefully setting weights. 

\begin{figure}[htbp!]
	\centering
	\includegraphics[width=1\linewidth]{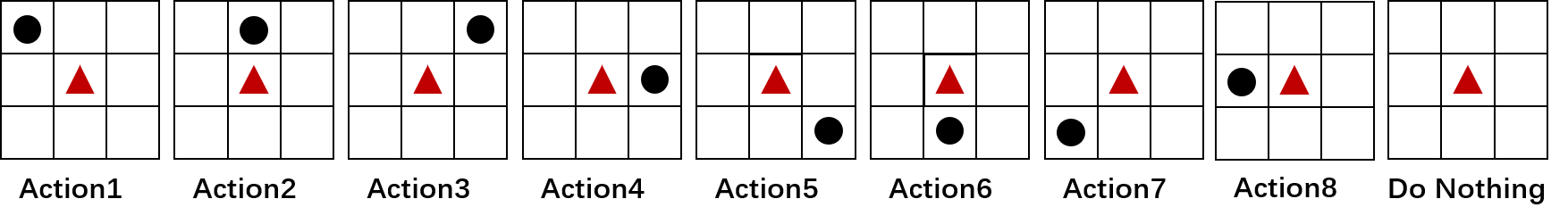}
	\caption{Sub-actions. The red triangle denotes the position corresponding to the agent and the black circle denotes the position of neighbor pixel.}
	\label{fig:Averaging actions}
\end{figure}

Consider the case of Gaussian denoising. For each noisy pixel $g_{x,y}$, we have $g_{x,y}=f_{x,y}+n_{x,y}$ with $n_{x,y}\sim\oN(0,\sigma^2)$. For a noisy pixel $g_{x,y}$, if we can find an similar adjacent pixel $g_{x',y'}$, which satisfy $f_{x,y}\approx f_{x',y'}$, then $g_{x,y}$ can be updated by their average as follows:
$$
g'(x,y)=ag_{x,y}+(1-a)g_{x',y'}\approx f_{x,y}+an_{x,y}+(1-a)n_{x',y'}.
$$
Let $n'_{x,y}=an_{x,y}+(1-a)n_{x',y'}$, we have $g'(x,y)=f_{x,y}+n'_{x,y}$. Since it is assumed that noises are independent in Gaussian denoising, we have $n'(x,y)\sim\oN(0,{[a^2+(1-a)^2]}\sigma^2)$, which means that by averaging $g_{x,y}$ and $g_{x',y'}$, the noise of $g_{x,y}$ decays with ratio $a^2+(1-a)^2$. The ratio $a^2+(1-a)^2$ get its minimum $0.5$ when $a=0.5$. Therefore, the noisy level can decay exponentially by averages, \ie~ the noise level can be reduced to $2^{-T}\sigma^2$ in $T$ steps. It should be noted that, the assumption that there exists an adjacent pixels similar with a pixel $f_{x,y}$, could be satisfied at most cases, \eg~a similar adjacent pixel on edge direction for an edge point, a similar adjacent pixel on an arbitrary direction for an textured pixels. Until now, we can define a simple but \XRT{sufficient} action sets as follows:
\begin{equation}\label{eq:action_set}
\sA_s=\{u^{(t+1)}_{x,y}\leftarrow 0.5u^{(t)}_{x,y}+0.5 u^{(t)}_{x+i,y+j}|(i,j)\in\sN\},
\end{equation}
The action set $\sA_s$ simply contains $8$ weighted average actions which averages the pixel value of the target pixel and its neighbor to eliminate the noise. Noted that each action in $\sA_s$ is an average operation, which may unavoidably blur the structures in the image after server decision steps. Therefore, to avoid over-smoothing, an "do nothing" action, which does not change that current state, is also added to the action set. The final action set is then defined as follows:
$$
\sA=\sA_s\cup\{u^{(t+1)}_{x,y}\leftarrow u^{(t)}_{x,y}\}
$$
The action set $\sA$ is illustrated in Fig.~\ref{fig:Averaging actions}. As the agents are supposed to choose the proper actions in each step, the proposed action set, though simple, can well perform diffusion along different directions in several steps, leading to a content-adaptive anisotropic diffusion model. 
\begin{remark}
	\textbf{Comparison with kernel prediction network:} the final denoised image $\vu^{(T)}$ can generated by performing spatially-variant weighted average on neighborhoods of the noisy image $\vg$, where the weights for each pixel can be compute according to the action perform in all the step. That is to say, the proposed diffusion model can be view as a new kernel prediction network. However, compared with often-seen kernel prediction networks, two main differences exist. 
	Firstly, the kernels of our diffusion model are generated step by step, while they are predicted directly in often-seen kernel prediction networks. The step-by-step scheme may bring some benefits as the intermediate denoised images, which have less noise than the noisy one, may helps to predict better kernels.
	Furthermore, unlike the often-seen kernel prediction networks, which predict kernel with regular shapes and limited sizes, the kernels generated by our diffusion model can have irregular shape. The irregular shape can be adaptive to the image contents, \eg~an edge-liked shape for an edge. In addition, the size of our kernels can goes to $2^T$ in $T$ steps.
\end{remark}
\section{DRL Framework}
\begin{figure}
	\includegraphics[width=\linewidth]{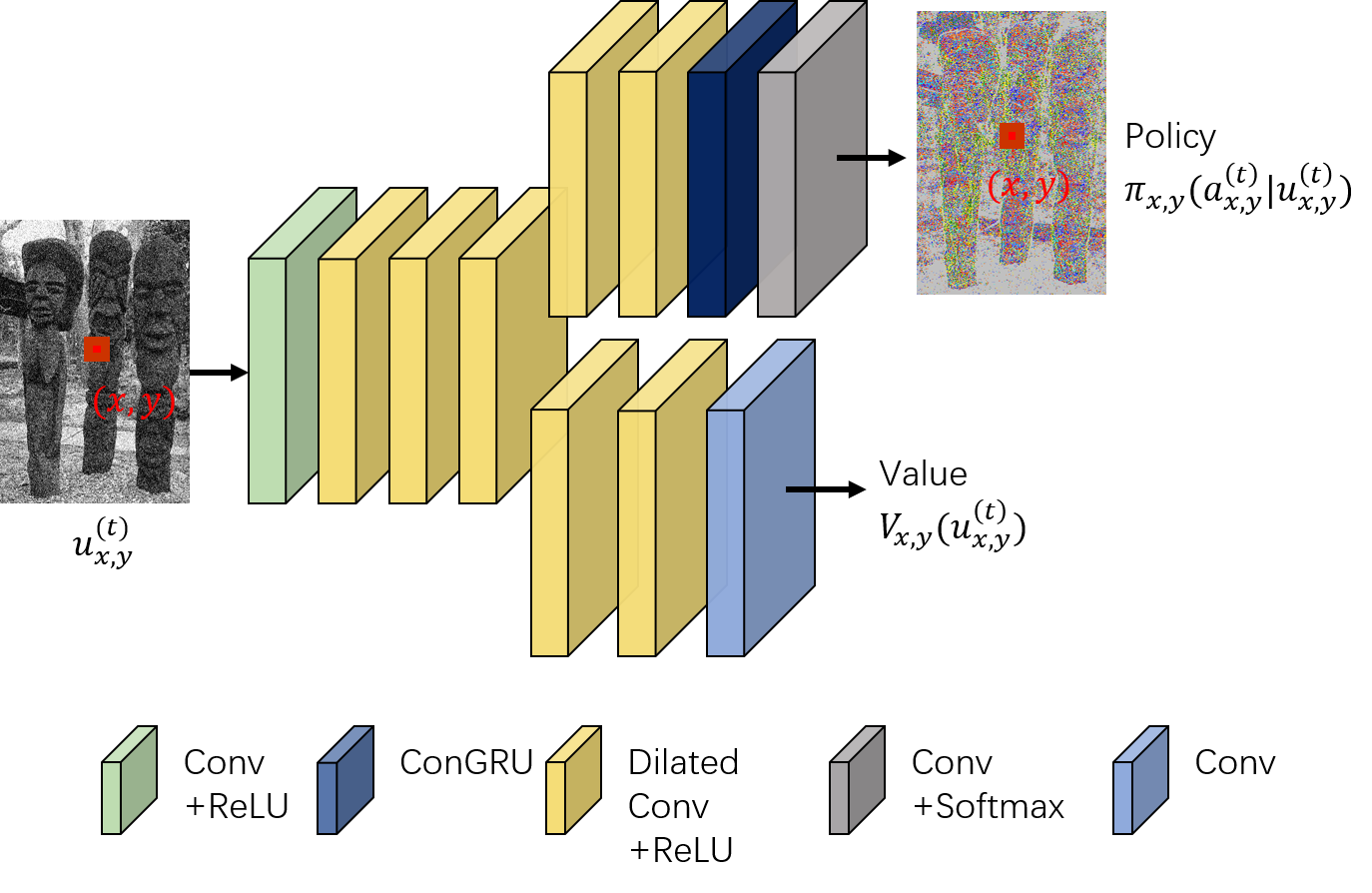}
	\caption{The network architecture of the deep reinforcement framework.\label{fig:architecture}}
\end{figure}

The MDP described above can be solved in a PixelRL framework~\cite{furuta2020pixelrl}, which extends the asynchronous advantage actor-critic (A3C) algorithm to an agent-per-pixel configuration. Following~\cite{furuta2020pixelrl}, we first define the return of the $(x,y)$-th agent at time $t$ as follows:
\begin{equation}\label{eq:conv_return}
G^{(t)}_{x,y}=r_{x,y}^{(t)}+\gamma\sum_{(i,j)\in\sN}\omega_{i,j}G_{x+i,y+j}^{(t+1)}.
\end{equation}
where $\sN$ is the local window centered at the $(x,y)$-th pixel and $\omega_{i,j}$ is the weight that means how much we consider the return of the neigbour pixels at the next time step. Unlike the original definition~\eqref{eq:return}, the definition \eqref{eq:conv_return} not only consider the return of the $(x,y)$-th agent at the next time step but also the return of the neighbor agents at the next time step. It is a desired property in our MDP since the action $a^{(t)}_{(x,y)}$ not only affects the $(x,y)$-th pixel in the next state, but also the neighbor pixels and their policies. By arrange $G^{(t)}_{x,y}$ into a two-dimensional space $\Omega$, the return of the whole MDP at time $t$ can be express as the following forms:
\begin{equation}\label{eq:conv_return_map}
\mG^{(t)}=\vr^{(t)}+\gamma\bm{\omega}*\mG^{(t+1)}=\sum_{k=0}^{\infty}\gamma^{n}\vw^{n}*\vr^{(t+k)},
\end{equation}
where * is the convolution operator, and $\bm{\omega}$ denotes the weights of the neighbouring pixels. 
Let $\bm{\pi}:\vs\rightarrow [0,1]^{|\sA|^{|\Omega|}}$ denote the policy of the whole MDP, then the value of state $\vs$ can be defined as
$$
	\mV^{\bm{\pi}}(\vs)=\sE_{\bm{\pi}}[\mG^{(t)}|\vs^{(t)}=\vs].
$$
The goal of the reinforcement learning is to  find a policy $\pi$ that maximizes the average of  discounted accumlated rewards in this MDP:
$$
\bm{\pi}^*=\argmax_{\bm{\pi}}~\sE_{\bm{\pi}}[\bar{\mG}^{(0)}],\text{~where~} \bar{\mG}^{(0)}=\frac{1}{\Omega}\sum_{(x,y)\in\Omega}G_{x,y}^{(0)}.
$$
Similar with the single-agent A3C, policy network $\bm{\pi}_{\rho}$ and value network $\mV_\theta$ are learned to approximate the optimal policies $\bm{\pi}^*(\vs,\va)$ and the state value function $\mV^{\bm{\pi}^*}(\vs^{(t)})$, respectively. The architectures of the both networks are illustrated in Fig.~\ref{fig:architecture}. Regarding the value network, its parameters $\theta$ are learned by iteratively minimizing a sequence of loss functions, where the loss function at time $t$ is defined as
\begin{equation}
\ell^{(t)}(\theta)=\norm{\vr^{(t)}+\gamma\bm{\omega}*\mV_{\theta'}(\vs^{(t+1)})-\mV_{\theta}(\vs^{(t)})}_F^2,
\end{equation}
where $\theta'$ is the parameters before update. 

Regarding the policy network, we first introduce the advantages of the agent-per-pixel MDP:
$$
	\mA(\va^{(t)}|\vs^{(t)})=\vr(\vs^{(t+1)})+\gamma \mV_\theta(\vs^{(t+1)})-\mV_\theta(\vs^{(t)}).
$$
Following A3C, the parameters $\rho$ of the policy network are learned by iteratively maximizing the average of the expected advantages:
\begin{equation}\label{eq:expected_adv}
\max_\rho\sE_{\bm{\pi}_\rho}[\bar{\mA}(\va^{(t)}|\vs^{(t)})]=\sum_{\va^{(t)}\in\sA^{|\Omega|}} \bm{\pi}_{\rho}(\va^{(t)}|\vs^{(t)})\bar{\mA}(\va^{(t)}|\vs^{(t)}),
\end{equation}
where $\bar{\mA}(\va^{(t)}|\vs^{(t)})$ denotes the average of advantages. In such a case, the output dimension of the policy network $\bm{\pi}_{\rho}(\va|\vs)$ is $|\sA|^{|\Omega|}$, which is extremely high and computationally impractical. To simplify, we assume the policies for different agents are independent, so that the problem~\eqref{eq:expected_adv} can be divided in to $|\Omega|$ independent subproblems as follows:
\begin{equation}\label{eq:expected_adv_sub}
\max_{\pi_{\rho_{x,y}}}\pi_{\rho_{x,y}}(a_{x,y}^{(t)}|\vs^{(t)})A_{x,y}(a_{x,y}^{(t)}|\vs^{(t)}).
\end{equation}
where $A_{x,y}(a_{x,y}^{(t)}|\vs^{(t)})$ is the $(x,y)$-th elements of $\mA(\va^{(t)}|\vs^{(t)})$ and $\pi_{\rho_{x,y}}$ is the policy for the $(x,y)$-th agent.
The problems can then be solved by $N$ networks with $|\Omega|$ outputs. However, it is still inefficient to train $N$ different networks. Considering that the policies should only depends on the the content of the neighbor pixels, we assume that the policy at each agent should be same, then the problem~\eqref{eq:expected_adv} can be rewritten as 
$$
	\max_{\pi_{\rho}}[\sum_{x,y\in\Omega}\pi_{\rho}(a_{x,y}^{(t)}|\vs^{(t)})A_{x,y}(a_{x,y}^{(t)}|\vs^{(t)})].
$$
In such a case, the policy network can be implemented with a Fully Convolutional Network, where the policies for each agent sharing the same parameters due to the convolution architectures. 

\section{Experiments}\label{sec:exp}


For evaluation, we apply our model to three denoising tasks: guassian denoising, possion denosing, salt and pepper noise removal.
For all the tasks, we used BSD68 dataset~\cite{martin2001database}, from which we use 428 images for training and 68 images for testing. In the training procedure, patches with size of $70\times70$ are randomly cropped from the images and fed into the proposed network.

In all experiments, we set $T=5$, $\gamma=0.95$. For the network training, we used ADAM~\cite{kingma2014adam} as the optimizer, and set the batch size to 64. We train the model for $Q=60000$ epochs in total. The learning rate is setting to $1.0\times10^{-3}$ at beginning and  decays with the factor $(1-\frac{i}{Q})$ after each episode. 
To reduce the training time, we initialize the weights of FCN with the publicly available weights of~\cite{zhang2017beyond}. Following~\cite{furuta2020pixelrl}, we also adopted a stage-wise training scheme: the FCN is trained firstly, and subsequently it was trained again with the reward map convolution.

Several representative/state-of-the-art approaches are selected for comparison, including BM3D~\cite{dabov2008image}, WNNM~\cite{gu2017weighted},  MLP\cite{burger2012image}, DnCNN ~\cite{zhang2017beyond}, PM~\cite{perona1990scale}, $4^\text{th}$ PDE~\cite{you2000fourth}, TNRD~\cite{chen2017trainable}  and PixelRL~\cite{furuta2020pixelrl}. Among these methods, BM3D and WNNM are two state-of-the-art non-learning denoisers, MLP and DnCNN are two state-of-the-art NN-based denoisers. PM, $4^\text{th}$ PDE and TNRD are three representative diffusion-based denoisers, where TNRD is also a learnable model. In comparison, the results of the best performed TNRD are reported,which is fully trained with $24$ filters of size $5\times 5$ and $5$ stages. Note that the number of stages is the same as the diffusion steps for fair comparison. PixelRL is a representative DRL-based method for image restoration. In addition, since training with data augmentation can further boost the performance of the proposed method, we also report the result produced by model trained with data augmentation in evaluation. The obtained model is denoted by "Ours+", whose data augmentation is done by left-right flips and rotations on the training images.
%
%
%
\subsection{Bahavior analysis}
\begin{figure*}
	\newcommand{\INPUT}{fig/action_map/0_INPUT.png}
	\newcommand{\OUTPUT}[1]{fig/action_map/0step#1_output.png}
	\newcommand{\ACTION}[1]{fig/action_map/image0_step#1.png}	
	\newcommand{\WIDTH}{0.15\linewidth}
	\centering
	\begin{minipage}{\WIDTH}
		\centering
		\includegraphics[width=\linewidth]{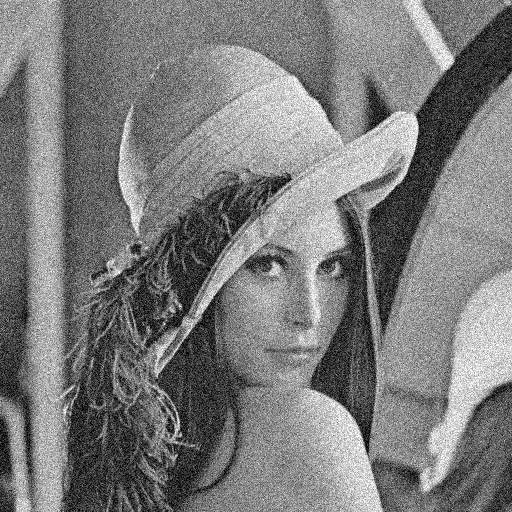}\\[2pt]
		\includegraphics[width=\linewidth,height=\linewidth]{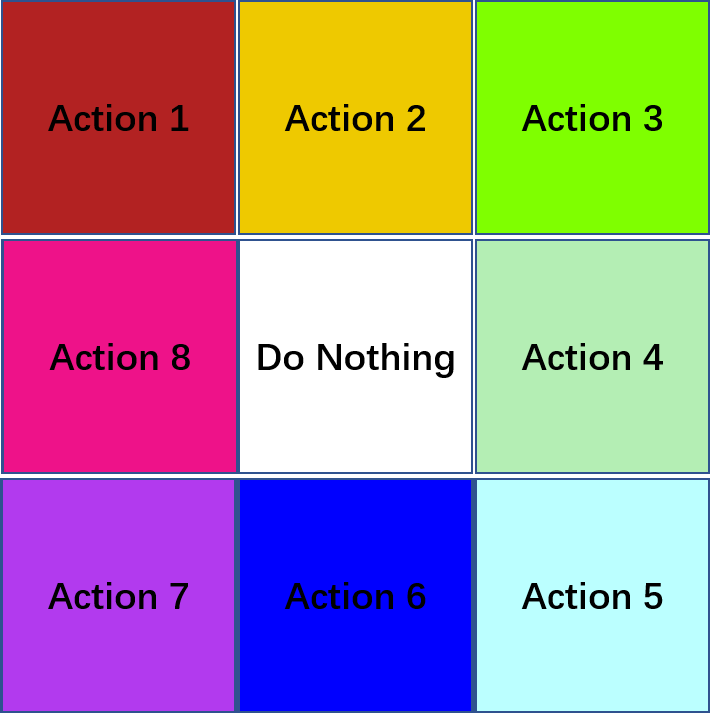}
		t=0
	\end{minipage}
	\newcommand{\vpair}[2]{
		\begin{minipage}{\WIDTH}
			\centering
			\includegraphics[width=\linewidth]{\OUTPUT{#2}}\\[2pt]
			\includegraphics[width=\linewidth]{\ACTION{#2}}
			t=#1
		\end{minipage}
	}
	\vpair{1}{0}
	\vpair{2}{1}
	\vpair{3}{2}
	\vpair{4}{3}
	\vpair{5}{4}
	\caption{Denoising process of the proposed method and the action map at each time step.\label{fig:action_map}}
\end{figure*}

We visualize the denoising process of the proposed method and the action map at each time step in Fig~\ref{fig:action_map}. It can be observed that the edges of the image are exhibited with white color in the action maps, which implies that the policy network can well extract the edges and "do nothing" actions can then be performed with high probabilities on extracted edges for edge preservation. Note that the edges in the action maps become stronger as the step grows. A probable reason is that as the noise is gradually reduced, the policy network can judge the edge with higher confidence, leading to more "do nothing" actions on the edges and stronger white edges in the action maps. Some other interesting phenomenons can be observed at the close-ups of the action maps. In the close-ups of a vertical edge, it can be seen that red(Action 1), purple~(Action 4), and pink~(Action 8) are the dominant colors at the left side of the edge, while green~(Action 3), light green~(Action 4) and light blue~(Action 5) are the dominant colors on the right side. It shows that the policy network tends to avoid diffusion across edges, which is expected behavior in diffusion-base denoising. Similar phenomenon can also be seen in the close-ups of the approximatedly horizontal edges, where blue~(Action 6) and purple~(Action 7) colors dominate the region below the edges.

\begin{figure}
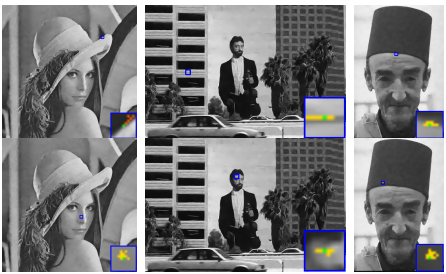

	\newcommand{\EDGE}[1]{fig/kernel/edge/Samecolor#1_output4point.png}
	\newcommand{\AREA}[1]{fig/kernel/area/Samecolor#1_output4point.png}
	\newcommand{\WIDTH}{0.2\linewidth}
	\centering
	\includegraphics[height=\WIDTH]{\EDGE{0}}
	\includegraphics[height=\WIDTH]{\EDGE{6}}
	\includegraphics[height=\WIDTH]{\EDGE{7}}\\
	\includegraphics[height=\WIDTH]{\AREA{0}}
	\includegraphics[height=\WIDTH]{\AREA{6}}
	\includegraphics[height=\WIDTH]{\AREA{7}}
	\caption{Visualization of the composited average operations.\label{fig:kernel}}
\end{figure}

Consider that only average operations are performed in the whole denoising process. Therefore, by compositing the averages in all the steps, the final denoised image can be generated by composited averages on the noisy image.  Here, we also show some examples of the composited average operations. See Fig.~\ref{fig:kernel} for an illustration. It can be seen that, for pixels on the edge~(the 1st row in the figure), the composited averages only involve pixels along the edges, while for pixels in a plain/textured regions, the averages involve the surrounding pixels, \XRT{which is also an expected behavior in edge-preserving denoising.}


\subsection{Results and Comparison}
\begin{table*}[htbp]
	\centering
	\newcommand{\PreserveBackslash}[1]{\let\temp=\\#1\let\\=\temp}
	\newcolumntype{C}{>{\PreserveBackslash\centering}p{0.6cm}}
	\newcolumntype{W}{>{\PreserveBackslash\centering}p{0.8cm}}
	\newcolumntype{K}{>{\PreserveBackslash\centering}p{1.0cm}}
	\begin{minipage}{\linewidth}
			\centering
			\caption{ Average PSNR(dB) of denoising results on BSD68 by different methods on Gaussian noise images. The best and second best results are highlighted in \textbf{bold} and \underline{\textit{underlined}} fonts.}
			\begin{tabular}{KCCCCCCCCCC}
			
			\toprule
			\multirow{2}[4]{*}{$\sigma$} & \multicolumn{10}{c}{Methods} \\
			\cmidrule{2-11}          & BM3D  & WNNM  & MLP   & DnCNN & PM    & $4^\text{th}$~PDE & TNRD  & PixelRL & Ours & Ours+ \\
			\midrule
			15    & 31.07 & 31.42 & -     & \textbf{31.63} & 25.20 & 24.49 & 31.30 & 31.40 & 31.48 & \textit{31.58} \\
			25    & 28.57 & 28.92 & 28.96 & \textbf{29.15} & 24.96 & 23.42 & 28.78 & 28.85 & 29.01 & \textit{29.10} \\
			50    & 25.62 & 25.97 & 26.03 & \textbf{26.19} & 23.06 & 20.90 & 25.80 & 25.88 & 26.08 & \textit{26.14} \\
			\bottomrule
			
			\end{tabular}%
			\label{tab:gaussian results}
	\end{minipage}\\[5pt]
	\begin{minipage}{0.49\linewidth}
		\centering
		\caption{ Average PSNR(dB) of denoising results on BSD68 on Salt and Pepper noise images. The best and second best results are highlighted in \textbf{bold} and \underline{\textit{underlined}} fonts.}
		\begin{tabular}{KCCCWWCC}	
			\toprule
			\multirow{2}{1.0cm}{\centering Noise  density(\%)} &       & \multicolumn{6}{c}{Methods} \\
			\cmidrule{2-8}         & BM3D  & DnCNN & PM    & 4th~PDE & PixelRL & Ours & Ours+	 \\
			\midrule
			10    & 24.61 & \textbf{40.16} & 21.96 & 20.91 & 38.46 & 39.34 & \textit{40.08} \\
			50    & 17.92 & 29.19 & 14.51 & 15.05 & 29.78 & \textit{31.4} & \textbf{31.65} \\
			90    & 13.5  & 23.58 & 10.70  & 11.77 & 23.78 & \textit{24.01} & \textbf{24.11} \\
			\bottomrule
		\end{tabular}%
		\label{tab:Salt and Pepper results}
	\end{minipage}\hspace{6pt}
	\begin{minipage}{0.49\linewidth}
		\centering
		\caption{ Average PSNR(dB) of denoising results on BSD68 on Poisson noise images.The best and second best results are highlighted in \textbf{bold} and \underline{\textit{underlined}} fonts.}
		\begin{tabular}{KCCCWWCC}
			\toprule
			\multirow{2}[4]{1.0cm}{\centering Peak Intensity} &       & \multicolumn{6}{c}{Methods} \\
			\cmidrule{2-8}          & BM3D  & DnCNN & PM    & 4th~PDE & \multicolumn{1}{l}{PixelRL} & \multicolumn{1}{l}{Ours} & Ours+ \\
			\midrule
			120   & 29.07 & \textbf{31.62} & 21.61 & 20.67 & 31.37 & 31.49 & \textit{31.58} \\
			30    & 24.84 & \textit{28.20} & 24.70 & 22.79 & 27.95 & 28.15 & \textbf{28.26} \\
			10    & 22.23 & \textbf{25.93} & 27.25 & 25.39 & 25.70 & 25.84 & \textit{25.91} \\
			\bottomrule
		\end{tabular}%
		\label{tab:Poisson results}
	\end{minipage}
\end{table*}%

\textbf{Gaussian Denoising:} in this subsection, we first evaluate the propsoed method on Gaussian denoising in three different levels: $\sigma= 15, 25, 50$. The results are shown in Table \ref{tab:gaussian results}. It can be seen that the proposed method achieves improvement with a large gap compared with two non-learning diffusion-based denoisers PM and $4^\text{th}$ PDE. The  Compared with another learnable diffusion model TNRD, the improvements are still obvious, which is $0.18,0.07,0.28$ for $\sigma=15,25,50$, respectively. On the other hand, the proposed method also showed improvements compared with another DRL-based denoiser PixelRL, which shows that the diffusion model can help to build a better deep reinforcement framework for image denoising. Furthermore, the performance of the proposed method can be further boosted with data augmentation. Our method with data augmentation ("Ours+") achieved the second best performance compared with all the other methods.

\textbf{Salt and pepper noise removal:} we also evaluate the proposed method for salt and pepper noise removal. Noises with density of $10\%$, $50\%$ and $90\%$ are added to clean images for evaluation. Since WNNM, MLP and TNRD are not designed for this task, we only compared with other methods in this experiments. The results are shown in Table~\ref{tab:Salt and Pepper results}.  It can be seen that the proposed method achieved a large gap of improvements compared with diffusion-based methods as well as another DRL-based method. With data augmentation, the proposed method can achieve the best results for density of $50\%$, $90\%$ and the second best result for density of $10\%$.

\textbf{Poisson denoising:} the performance of the proposed method on Poisson denoising is also evaluated. In this experiments, Poisson noise with peak of $120$, $20$, $10$ are imposed onto clean images for evaluation. Similarly, WNNM, MLP and TNRD are not compared as they are only designed for gaussian denoising. The results are shown in Table~\ref{tab:Poisson results}. It can be observed that the proposed method outperforms two diffusion-base methods and PixelRL, and achieve comparative results with the best one when using data augmentation.

\section{Conclusion and future work}\label{summary}
In this paper, we have proposed a learning-based diffusion model for image denoising, where the process of diffusion is determined by pixel-wise agents learned from a multi-agent deep reinforcement network. The proposed method has shown superior performance over existing learning-based diffusion model. In the future work, we are interested in extending the proposde method to a self-supervised setting, by designing a self-supervised deep reinforcement learning framework for the learning-based diffusion model. We are also interested in extending the learning-based diffusion model for other applications such as edge detection, texture removal.

\ifCLASSOPTIONcaptionsoff
  \newpage
\fi


\bibliographystyle{IEEETran}
\bibliography{DRD_v1}

%
%
%
%

\end{document}